\documentclass[10pt,twocolumn,letterpaper]{article}

\usepackage{iccv}
\usepackage{times}
\usepackage{epsfig}
\usepackage{graphicx}
\usepackage{amsmath}
\usepackage{amssymb}

\usepackage{bbm}
\usepackage{multirow}

\usepackage[pagebackref=true,breaklinks=true,letterpaper=true,colorlinks,bookmarks=false]{hyperref}
\iccvfinalcopy 
\def\iccvPaperID{8}

\DeclareMathOperator*{\argmin}{arg\,min}  

\begin{document}


\title{3D Shapes Local Geometry Codes Learning with SDF}

\author{Shun Yao$^{1,2}$, Fei Yang$^{1,2}$, Yongmei Cheng$^{1,2}$, Mikhail G. Mozerov$^{2}$\\
$^{1}$ School of Automation, Northwestern Polytechnical University, Xi'an, China\\
$^{2}$ Computer Vision Center, Universitat Autonoma de Barcelona, Barcelona, Spain\\
{\tt\small \{shunyao, fyang, mozerov\}@cvc.uab.cat, chengym@nwpu.edu.cn}
}

\maketitle
\ificcvfinal\thispagestyle{empty}\fi

\begin{abstract}

A signed distance function (SDF) as the 3D shape description is one of the most effective approaches to represent 3D geometry for rendering and reconstruction. Our work is inspired by the state-of-the-art method DeepSDF~\cite{park2019deepsdf} that learns and analyzes the 3D shape as the iso-surface of its shell and this method has shown promising results especially in the 3D shape reconstruction and compression domain.
In this paper, we consider the degeneration problem of reconstruction coming from the capacity decrease of the DeepSDF model, which approximates the SDF with a neural network and a single latent code.
We propose Local Geometry Code Learning (LGCL), a model that improves the original DeepSDF results by learning from a local shape geometry of the full 3D shape.
We add an extra graph neural network to split the single transmittable latent code into a set of local latent codes distributed on the 3D shape.
Mentioned latent codes are used to approximate the SDF in their local regions, which will alleviate the complexity of the approximation compared to the original DeepSDF.
Furthermore, we introduce a new geometric loss function to facilitate the training of these local latent codes.
Note that other local shape adjusting methods use the 3D voxel representation, which in turn is a problem highly difficult to solve or even is insolvable. In contrast, our architecture is based on graph processing implicitly and performs the learning regression process directly in the latent code space, thus make the proposed architecture more flexible and also simple for realization.
Our experiments on 3D shape reconstruction demonstrate that our LGCL method can keep more details with a significantly smaller size of the SDF decoder and outperforms considerably the original DeepSDF method under the most important quantitative metrics.
   
\end{abstract}

\section{Introduction}

Recently, deep neural networks is used to model implicit representations of 3D shapes has been widely applied for reconstruction \cite{duggal2021secrets}, generation \cite{chen2019learning}, compression \cite{tang2020deep} and rendering \cite{remelli2020meshsdf, takikawa2021neural}.
As one of the most popular methods, DeepSDF~\cite{park2019deepsdf} represents the zero-level surface of the whole 3D shape by regressing its continuous signed distance function (SDF).
However, the effectiveness of such models depends on the complexity of 3D shapes and the capacity of neural networks.
In Fig~\ref{fig:compare2degeneration}~(d) compared to (c) we demonstrate an example, where the capacity of the model is insufficient. Consequently,  the reconstruction of one complex 3D shape is of poor quality.

To alleviate this problem, we propose to learn a set of local SDFs to represent the whole surface. In this case,  each local SDF is responsible for a part of the reconstructed shape.
Learning such local SDFs is much easier. Here we use the assumption that the complexity of one local part of the 3D shape is much simpler than the whole and usually similar to other local parts.

However, learning a set of local SDFs instead of one global SDF makes the training process of the model more difficult, since the distribution of training data is not uniform, especially when more SDFs are considered in a local region.
One of the possible solutions is to make these local SDFs learnable using both the database and the local latent code. Here we utilize the reasonable assumption that similarity of the neighbor parts means similarity in the local code of the shape.

Recently, graph neural networks (GNNs) have demonstrated high effectiveness in geometric learning field, especially in 3D reconstruction problem~\cite{ranjan2018generating,bouritsas2019neural,zhou2020fully,hanocka2019meshcnn}.
The discussed works show that it is possible to decode latent representation of the shape geometry with highly accurate reconstruction, even for finer details.

The GNNs provide geometric constraints to facilitate the information smoothed over the graph as the original purpose for clustering vertices of a graph~\cite{kipf2016semi, defferrard2016convolutional}.
This property leads to a message exchange mechanism among the vertices and results in their local similarity.

Consequently, we introduce Global-to-Local~(G2L) network based on GNNs to learn local latent codes that came from one global code. Thus, our G2L combines the advantages of the GNNs and SDF approaches.
We assume that geometrical locality is learnable in the space of latent code as in the original geometric space of the shape.

In addition, we also include one geometric similarity loss function based on the geometric structure to enhance the effectiveness of the GNN in local latent code learning.
Remind our assumption is that the neighboring local regions should have similar latent codes in general. Our experimental results confirm the mentioned assumption with the local latent codes learning.

Several methods~\cite{chabra2020deep, jiang2020local, genova2020local, tretschk2020patchnets, hao2020dualsdf}  learn the SDF of 3D shape locally and have shown promising results.
However, all of them need either the voxel representation of 3D shape to align the local latent codes or explicit parametric models (for example, a sphere) to fit.
Both approaches face problems of the volumetric representation that in many cases highly difficult to solve or even is insolvable.
In contrast, our work leverages geometric learning techniques to model local SDFs of 3D shapes directly in the latent code space.

 Contributions of our work are:
\begin{itemize}
    \item We propose the Local Geometry Code Learning method, where we learn the shape as zero-surfaces with local latent codes.
    \item We use graph neural networks to generate local latent codes and distribute them on the 3D shape, which does not request voxelization of the 3D shape as it is in locally modeling methods.
    \item We introduce a geometric similarity in the loss function that helps to learn and reduce the fluctuation of the reconstructed surface.
    \item Our experimental validation shows that the proposed approach could keep more details of the reconstructed shape in comparison with the original SDF decoder.
\end{itemize}

\begin{figure*}
\begin{center}
\includegraphics[width=0.95\linewidth]{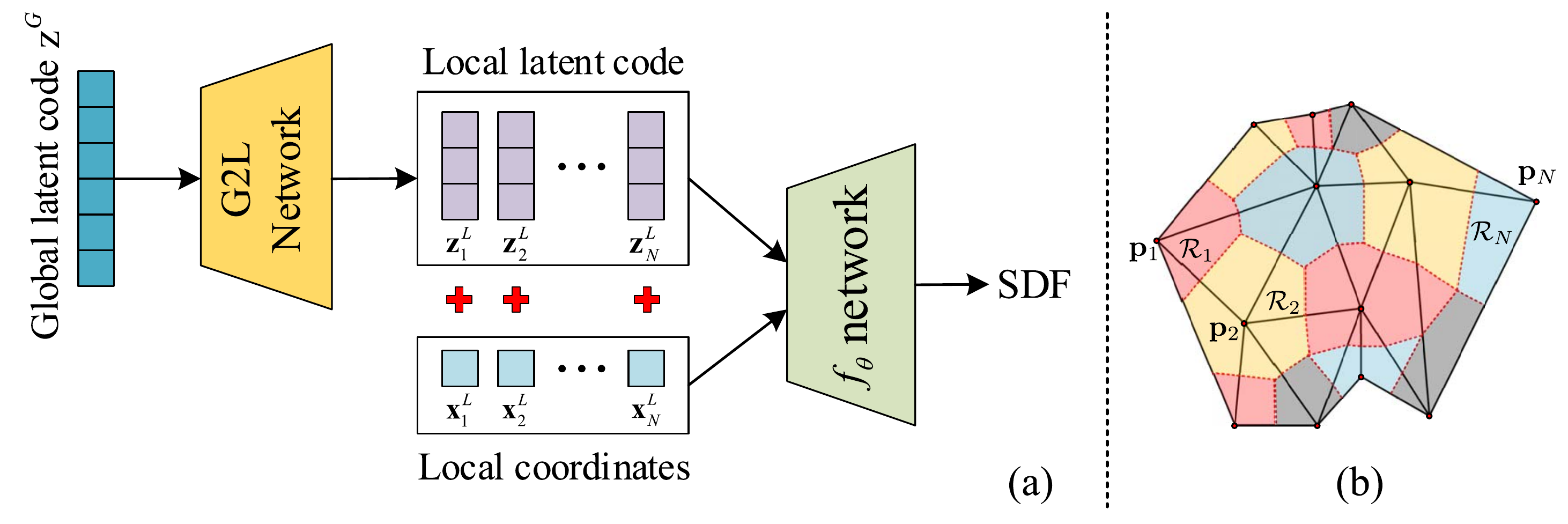}
\end{center}
\caption{(a) Overview of the  LGCL model; (b) Local region separation, where one 3D shape sample is represented with mesh data and projected to a 2D plane.}
\label{fig:geo_sdf_decoder}
\end{figure*}

\begin{figure}
\begin{center}
\includegraphics[width=0.9\linewidth]{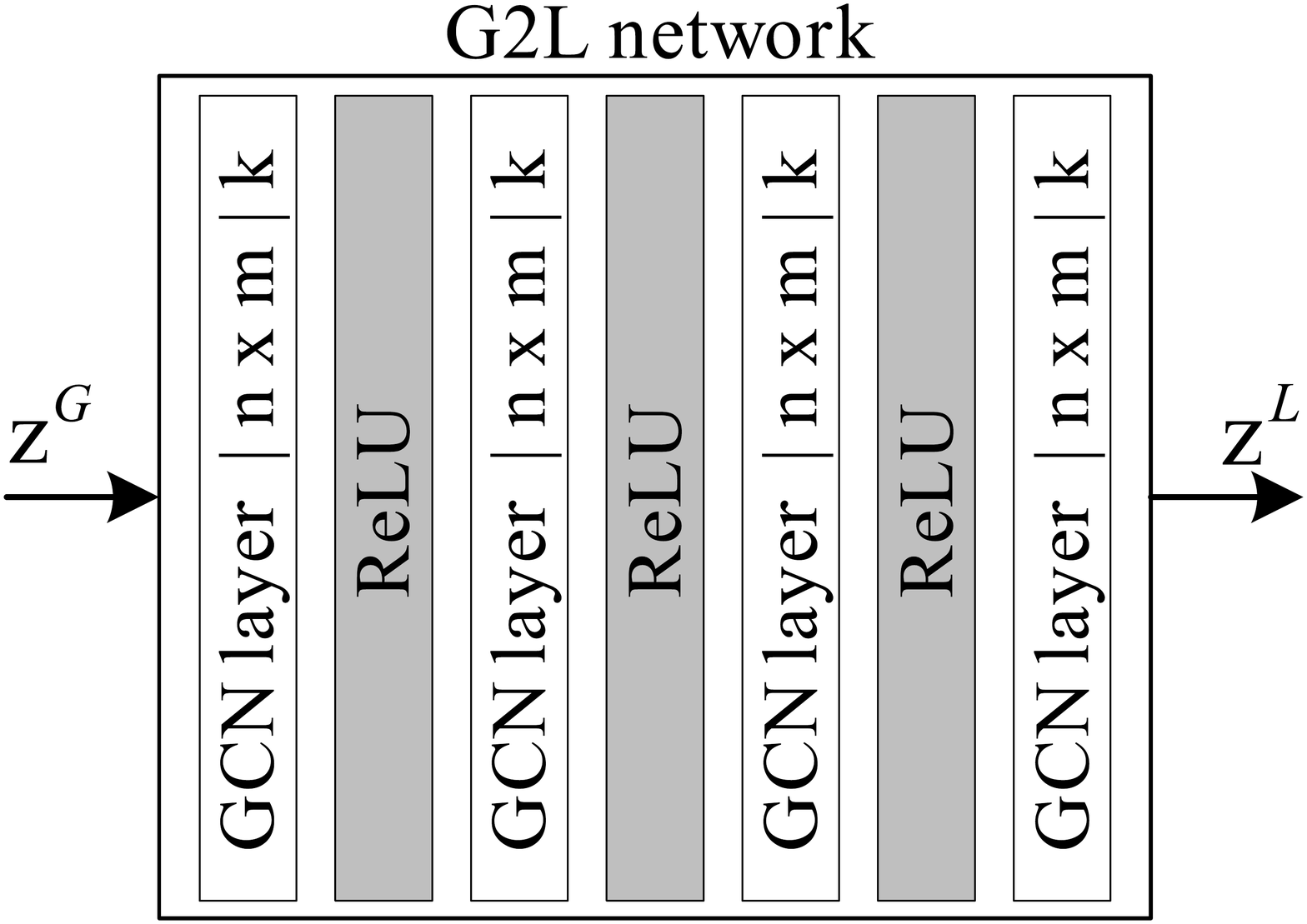}
\end{center}
\caption{Architecture of the G2L network.}
\label{fig:g2l}
\end{figure}

\section{Related work}
\subsection{SDF learning methods}

Learning a signed distance function to represent the 3D shape as a set of iso-surfaces recently receives extensive attention in the field.
Chen and Zhang~\cite{chen2019learning} proposed to assign a value to each point in 3D space and use a binary classifier to extract an iso-surface.
Mescheder~\etal~\cite{mescheder2019occupancy} utilize a truncated SDF to decide the continuous boundary of 3D shapes. In contrast with~\cite{chen2019learning}, they predict the probability of occupancy in voxels, which could be used in a progressive multi-resolution procedure to get refined output.
Park \etal~\cite{park2019deepsdf} learn a continuous field of 3D shape instead of the discrete representation of SDF in the grid and is understandable as a learned shape-conditioned classifier for the decision boundary of 3D shape surface.

Atzmon and Lipman~\cite{atzmon2020sal} leveraged sign agnostic learning (SAL) to learn from unsigned distance data, such as the distance directly from points to triangle soups.
Gropp \etal~\cite{gropp2020implicit} suggested using a geometric regularization paradigm to approximate the signed distance function, which can be achieved without 3D supervision and/or a direct loss on the surface of the shape.
Inspired by incorporating derivatives in a regression loss leads to a lower sample complexity, Atzmon and Lipman~\cite{atzmon2020sald} generalized SAL to include derivatives and show a significant improvement in the quality of 3D shape reconstruction.

Note that it is extremely hard to obtain the ground-truth data about the signed distance from sampling points to the surface of 3D shape for training.
This is the main motivation for us to propose a method that can avoid such a voxalization used in algorithms that we discuss in the above paragraphs.

\subsection{Modelling local SDFs}

Instead of learning a single  SDF for representing a whole shape, Jiang \etal~\cite{jiang2020local} designed an embedding of local crops of 3D shapes during training, and optimize a set of latent codes on a regular grid of overlapping crops with one single shared decoder when run on inference.

Inspired by DeepSDF~\cite{park2019deepsdf}, Chabra \etal~\cite{chabra2020deep} replaced the dense volumetric SDF representation used in traditional surface reconstruction with a set of locally learned continuous SDFs defined by a single parameterized neural network.
In contrast with the voxels(grid)-based representation of SDFs, Genova \etal~\cite{genova2019learning} proposed a network to encode shapes into structured implicit functions (SIF) as a composition of local shape elements.

Tretschk \etal~\cite{tretschk2020patchnets} designed an explicit parametric surface model, which fits an implicit SDF in each local patch separated from one sphere.

Hao \etal~\cite{hao2020dualsdf} represent 3D shapes as two levels of granularity with SDF, which provides interpretability for latent space of SDF in local parts of the shape.
They introduced a novel shape manipulation method by editing the primitives of local SDFs.

In the most recent works, Genova \etal~\cite{genova2020local} developed the SIF to learn a set of local SDFs that are arranged and blended according to a SIF template.
The method associates a latent vector with each local region that can be decoded with the SDF to produce finer geometric detail.

\subsection{Geometric learning on 3D shapes}
Generalize neural networks to data with the non-Euclidean structure are known as Graph Neural Networks (GNNs) in the domain of geometric learning.
Ranjan \etal~\cite{ranjan2018generating} proposed to learn a non-linear representation of human faces by spectral convolutions with Chebychev biasis~\cite{kipf2016semi, defferrard2016convolutional} as filters.
Bouritsas \etal~\cite{bouritsas2019neural} replaced the convolution kernel with operators applied along a spiral path around the graph vertices.

Hanocka \etal~\cite{hanocka2019meshcnn} leveraged the intrinsic geodesic connections of edges to define convolution operators, which inherited the direction invariant property as in 3D points convolution methods~\cite{qi2017pointnet, qi2017pointnet++}.

Zhou \etal~\cite{zhou2020fully} further improved the reconstruction precision by using locally adaptive convolution operators for registered mesh data.

We incorporate several ideas of this and the previous subsection in our method, but do it in different ways. More details are in the next section below.

\section{Local geometry code learning method}

\subsection{Modeling SDF locally}

A shape $\mathcal{S}$ can be represented as the zero level-set of $f_{\theta}(\mathbf{x}, \mathbf{z})$ as:
\begin{equation}
    \mathcal{S} = \{ \mathbf{x} \in \mathbb{R}^3 \mid f_{\theta}(\mathbf{x,z})=0 \},
\end{equation}
where $f_{\theta}(\mathbf{x},\mathbf{z}):\mathbb{R}^3 \times \mathbb{R}^m \rightarrow \mathbb{R}$ is a signed surface distance function implemented as a neural network (usually as multilayer perceptron network) with learnable parameters $\theta$. The latent code $\mathbf{z}$ decides output shape of $f_{\theta}$ along with the sampling coordinates $\mathbf{x}$.

Similar with DeepLS~\cite{chabra2020deep}, we want to make $f_{\theta}$ model the whole shape as a composition of its local parts:
\begin{equation}
    \mathcal{S} = \{ \mathbf{x} \in \mathbb{R}^3 \mid \bigcup\nolimits_{i} \mathbbm{1}_{\mathbf{x} \in \mathcal{R}_i} f_{\theta}(T_i(\mathbf{x}),\mathbf{z}_{i}^{L})=0 \},
\end{equation}
where $T_i(\cdot)$ supposes to transfer global location $\mathbf{x}$ to the local coordinate system $\mathbf{x}_{i}^{L}$ of a local region $\mathcal{R}_i$, and $\mathbf{z}_i^{L}$ indicates its related local latent code, as illustrated in Fig~\ref{fig:geo_sdf_decoder}~(a).

Different from splitting the 3D space into volumes~\cite{jiang2020local,chabra2020deep} or explicitly parametric surface model~\cite{tretschk2020patchnets, hao2020dualsdf}. we define the local region $\mathcal{R}_i$ with a key point $\mathrm{p}_i$ as:
\begin{equation}
    \mathcal{R}_i = \{ \mathbf{x} \in \mathbb{R}^3 \mid  \argmin\nolimits_{\mathbf{p} \in \mathcal{P}} d(\mathbf{x}, \mathbf{p}) = \mathbf{p}_i\},
\end{equation}
where $\mathcal{P}$ is a set of key points and $d(\cdot)$ is a distance function, \eg Euclidean distance.

Note that key points $\mathcal{P}$ are only used for aligning the sampling points to their corresponding local latent code $\mathbf{z}_i^L$. Thus,these key points do not necessary in the training or inference.

One simple illustration for our region division method is shown Fig~\ref{fig:geo_sdf_decoder}~(b). One patch with different color from their neighbours indicates a local region, which owns a corresponding latent code.

\subsection{Geometric leaning on local latent codes}

Different from the volume representation in DeepLS~\cite{chabra2020deep} and LDIF~\cite{genova2020local}, we do not know which local region includes the 3D shape, and it will lead to defining some unnecessary local latent codes.

Another problem is that these local latent codes are learned independently, which will lead to inconsistent surface estimates at the local region boundaries as mentioned in~\cite{chabra2020deep}.

Inspired by the geometric learning of COMA~\cite{ranjan2018generating}, Neural3DMM~\cite{bouritsas2019neural}, FCM~\cite{zhou2020fully}, we introduce the mesh structure of 3D shape as a "scaffold" to put key points and propagate information between local latent codes.
Then we can get two benefits that it allows to keep each local region has an intersection with the 3D shape and construct communications among them by representing the mesh as a graph.

Let us assume that a 3D surface mesh can be defined as a set of vertices $\mathcal{V}$ and edges $\mathcal{E}$, where $\mathcal{V}$ is replaced by $\mathcal{P}$ to define the local regions as shown in Fig \ref{fig:geo_sdf_decoder} (b).

Cooperate with several graph convolution layers to construct a graph network $\mathrm{G2L}(\mathcal{E}, \mathbf{z}^G): \mathbf{z}^G \rightarrow \{ \mathbf{z}^L_i \}$, we could predict the local SDF with the local latent codes aligned to these local regions, as shown in Fig \ref{fig:geo_sdf_decoder} (a).
Such graph neural network provides geometric deformations on each local latent code.
Consequently, each local latent code can contribute to the shape representation since each key point is on the shape.

Since our method gets benefits from modeling SDF in \textbf{Local} with latent codes and learning these latent codes through \textbf{Geometric Learning} with graph neural networks, thus we name it Local Geometry Code Learning (LGCL).

\subsection{Loss function}

\noindent {\bf Sign agnostic learning.}
Due to the advanced works of SAL~\cite{atzmon2020sal} and IGR~\cite{gropp2020implicit}, we do not request the true distance of sampling points to the surface of shape during training.
Instead of getting the true distance in a rendering way, directly calculating the distance from a point to a triangle soup is more convenient and efficient, and also without the requirement of watertight structures.

Thus, we construct basic loss function as:
\begin{equation}
    \mathcal{L}_{\mathrm{basic}} = \mathcal{L}_{\mathrm{sal}} + \mathcal{L}_{\mathrm{igr}} ,
\end{equation}
where $\mathcal{L}_{\mathrm{sal}}$ just needs the unsigned distance $d_{\mathrm{u}}$ from point $\mathbf{x}$ that sampled from whole space $\Omega$ to the shape $\mathcal{S}$, and it is defined as:
\begin{equation}
    \mathcal{L}_{\mathrm{sal}} = \mathbb{E}_{\mathbf{x}\in\Omega} \big| \left|f_{\theta}(T_i(\mathbf{x}), \mathbf{z}_i^L) \right| - d_{\mathrm{u}}(\mathbf{x}, \mathcal{S}) \big|.
\end{equation}
For the $\mathcal{L}_{\mathrm{igr}}$, we use its variant type from Siren~\cite{sitzmann2020implicit} as:
\begin{equation}
\begin{aligned}
    \mathcal{L}_{\mathrm{igr}} &= \lambda_{\mathrm{grad}} \mathbb{E}_{\mathbf{x}\in\Omega}(\| \nabla_{\mathbf{x}}f_{\theta} \|_2 -1)^2 +\\
    &\quad \quad
    \mathbb{E}_{\mathbf{x}\in\Omega_0} \| \nabla_{\mathbf{x}}f_{\theta} - n(\mathbf{x})  \|_2,
\end{aligned}
\end{equation}
where $\Omega_0$ means the domain of zero-iso surface of the shape, $\|\cdot\|_2$ is the euclidean 2-norm.

\noindent {\bf Geometric similarity loss.}
There is a contradiction between the distributions of key points and sampling points.
Consider a complex part of the shape, it needs more key points to get more local latent codes for better modeling.
However, the more key points are allocated, the harder the optimization of local latent codes is.
Since each local region would be smaller and get fewer sampling points for training.
Even if increasing the number of sampling points, it is still difficult to ensure assigning enough sampling points to each local latent code.

To alleviate this problem, we propose a loss $\mathcal{L}_{\mathrm{sim}}$ to make these local latent codes not only learn from the sampling points but also learn from each other.
The assumption here is the difference between the adjacent local latent codes is smaller than the ones that are far away from each other.
On the other hand, it provides a kind of regularization effect on the local latent codes that cannot get sufficient training, which forces them to be similar to their neighbors.

Again, we use the geometric structure as a graph to calculate $\mathcal{L}_{\mathrm{sim}}$ as:
\begin{equation}
    \mathcal{L}_{\mathrm{sim}} = \frac{1}{N_{\mathrm{v}}} \sum^{N_{\mathrm{v}}}_{i} \left| \mathbf{z}_i^L -  \sum^{K}_{k} G_{\mathrm{l}}(\mathbf{z}_i^L, \mathcal{N}_k(i)) \right|
\end{equation}
where $G(x_i, \mathcal{N}(i)): x = \frac{1}{\left| \mathcal{N}(i) \right|} \sum_{j\in \mathcal{N}(i)}(x_j)$ means to update the value of $x_i$ by the average value of its neighbours.
Here $\mathcal{N}_k(i)$ means the neighbours of vertex $i$ in the $k$ layer.
And for better learning, we increase the neighbor region of one local latent code by $K$ layers.
We use $K=3$ layers for our experiments.

\noindent {\bf Total loss}
Our final loss function consists of above losses and a regular term $\| \mathrm{z}^G\|$ as:
\begin{equation}
    \mathcal{L}_{\mathrm{total}} = \lambda_{\mathrm{sim}} \mathcal{L}_{\mathrm{sim}} + \mathcal{L}_{\mathrm{basic}} + \lambda_{reg} \|\mathrm{z}^G\|_2
\end{equation}

In our experiments, we use the setting of $\lambda_{\mathrm{grad}}=0.1$, $\lambda_{\mathrm{sim}}=1.0$ and $\lambda_{\mathrm{reg}}=0.001$ if without extra explanation.
\section{Experiments}
In our experiments, all of the used models are trained and evaluated mainly on a subset of the D-Faust dataset~\cite{bogo2017dynamic}, which is the No.50002 subset of mesh registrations for 14 different actions about the human body, such as leg and arm raises, jumps, etc.
Due to the low variation between adjacent scans, we sample the used dataset at a ratio of 1:10 and then split them randomly with 90\% for training and 10\% for the test.

Our data pre-processing method inherits from both IGR~\cite{gropp2020implicit} and SAL~\cite{atzmon2020sal}, which will generate 600K sampling points from each object, 300K are sampled on the object surface with their normals and the other 300K are sampled from two Gaussian distributions centered at points on the object surface.

\subsection{Reconstruction}

As one of the main baselines, we train the DeepSDF with the same setting in ~\cite{park2019deepsdf} on the completed No.50002 sub dataset of the D-FAUST dataset.

In addition, we also train other two different sizes of DeepSDFs with the loss function $\mathcal{L}_{\mathrm{basic}}$: SDF-8, which is similar to the original DeepSDF, but with 8 fully connected layers and 512 neurons in each hidden layer. The dimension of its latent code is 256. One skip connection is also used at the fourth layer with the latent code. SDF-4, has 4 fully connected layers, 128 neurons in each one and none skip connection, and the length of latent code is 64. Each fully connected layer in both SDF-8 and SDF-4 except the last one is followed by Softplus activation and initialized as in ~\cite{atzmon2020sal}.

In our LGCL-based method, we use a 4-layers G2L (as shown in Fig.~\ref{fig:g2l}) network followed by an SDF-4 network. The graph convolution kernels in the G2L are from~\cite{ranjan2018generating} as chebConv or~\cite{zhou2020fully} as vcConv, we gave the difference of their performance in the following results.
we re-implemented the kernels in a more compact form, and the parameters of one graph convolution layer can be represented as $(\mathrm{n},\mathrm{m},\mathrm{k})$, where $\mathrm{n}$ means the size of input channel and $\mathrm{m}$ is the size of the output channel. $\mathrm{k}$ stands for the size of the Chebyshev filter when using chebConv and the number of weight basis when using vcConv.

More details about the architecture of our models can be viewed in Table~\ref{tab:arch}.
The local latent code for each vertex of the G2L is set to an 8 length vector.

\begin{table}
\begin{center}
\begin{tabular}{l|c|c}
\hline
& LGCL-VC & LGCL-Cheb \\
\hline
\multirow{4}{*}{G2L} & vcConv(8,8,8) & chebConv(8,8,6) \\
& vcConv(8,16,16) & chebConv(8,16,6) \\
& vcConv(16,32,32) & chebConv(16,32,6) \\
& vcConv(32,64,64) & chebConv(32,64,6) \\
\hline
\multirow{4}{*}{SDF-4}
& \multicolumn{2}{c}{Linear(67,128)} \\
& \multicolumn{2}{c}{Linear(128,128)} \\
& \multicolumn{2}{c}{Linear(128,128)} \\
& \multicolumn{2}{c}{Linear(128,1)} \\
\hline
\end{tabular}
\end{center}
\caption{Setting of the architecture in the LGCL method.  LGCL-VC means using the graph convolution kernels from~\cite{zhou2020fully} and LGCL-Cheb is from~\cite{ranjan2018generating}.}
\label{tab:arch}
\end{table}

We train all the models with 300 epochs at a learning rate of 5e-4 for the parameters of neural networks and 1e-3 for latent codes optimization.
Both learning rates are decayed to half after 200 epochs.
We evaluate all the methods on the split test dataset.
As same as in DeepSDF~\cite{park2019deepsdf}, the latent code will be estimated with the frozen neural network before the inference.

\begin{table}
\begin{center}
\setlength{\tabcolsep}{2pt}{
\begin{tabular}{l|c|c|c|c|c|c}
\hline
\multirow{2}{*}{Model} & Net & Latent & \multirow{2}{*}{CD} & \multirow{2}{*}{HD} & \multicolumn{2}{c}{ED} \\
 & Params & Params &  &  & Mean & Std \\
\hline
DeepSDF & 1.84 M & 0.26 K & 0.28 & 68.11 & 12.51 & 16.37 \\
SDF-8 & 1.58 M & 0.26 K & 0.22 & 59.86 & 6.94 & 10.30 \\
SDF-4 & 41.86 K & 0.06 K & 2.20 & 119.56 & 24.45 & 31.19 \\
LGCL-Cheb & 58.42 K & 55.12 K & 2.56  & 108.58 & 2.51 & 1.87 \\
LGCL-VC & 0.19 M & 55.12 K & 1.55 & 71.67 & 3.35 & 2.44 \\
\hline
\end{tabular}}
\end{center}
\caption{Quantitative evaluation. CD means Chamfer Distance, HD means Hausdorff Distance and ED stands for Euclidean Distance of the point to surface.
All the distances are represented in millimeters.
We directly run the DeepSDF code as the baseline.}
\label{tab:qe}
\end{table}

\begin{table}
\begin{center}
\setlength{\tabcolsep}{1pt}{
\begin{tabular}{l|c|c|c|c|c}
\hline
 \multirow{2}{*}{Model} & \multicolumn{2}{|c}{Error(mm)} & \multicolumn{3}{|c}{Percentage($\%$)}\\
 & $<50\%$ & $<90\%$ & $>5$ mm & $>10$ mm & $>20$ mm\\
\hline
DeepSDF & 6.19 & 37.79 & 57.42 & 33.27 & 17.44 \\
SDF-8 & 3.55 & 18.59 & 34.79 & 14.14 & 6.39 \\
SDF-4 & 11.91 & 73.42 & 34.80 & 43.07 & 25.37 \\
LGCL-Cheb & 2.13  & 5.21 & 11.44 & 0.03 & 0.00 \\
LGCL-VC & 2.92  & 6.85 & 24.47 & 0.98 & 0.00 \\
\hline
\end{tabular}}
\end{center}
\caption{Statistics of reconstruction errors.}
\label{tab:statis}
\end{table}

To evaluate the performance of reconstruction, we measure the Euclidean Distance (ED) from the vertices of ground truth to the surface of reconstruction generated from different methods.
We also report our results under the metrics of Chamfer Distance (CD) and Hausdorff Distance (HD).

Due to CD and HD are applied on the point cloud, then we sample 30000 points on both surfaces of ground truths and reconstruction for it.
For a more fair comparison, we also list the size of network parameters and latent codes of different methods, while both are necessary to represent 3D shapes.
All of these quantitative results are shown in Table~\ref{tab:qe}.

As one can see that SDF-only-based methods need much more parameters of network to achieve comparable performance with our method.
We can see that there is a positive correlation between the size of the DeepSDF network and its performance on reconstruction, as all the quantitative results of SDF-4 are worse than SDF-8's.

By introducing local latent codes, our LGCL-based model outperforms SDF-4 by approximately one order of magnitude under the metric of Euclidean distance.
Even compared to the DeepSDF-8 which has a huge size of parameters, our results still has competitive advantages and obtain the smallest Euclidean error as $2.51 \pm 1.87$ mm with chebConv kernels.

More details about the Euclidean errors of different methods can be found in Table~\ref{tab:statis}.
Consequently, LGCL-Vc decreases the errors of CD and HD of SDF-4 by about 30\% and 40\% respectively.

\begin{figure}[ht]
\begin{center}
\includegraphics[width=0.9\linewidth]{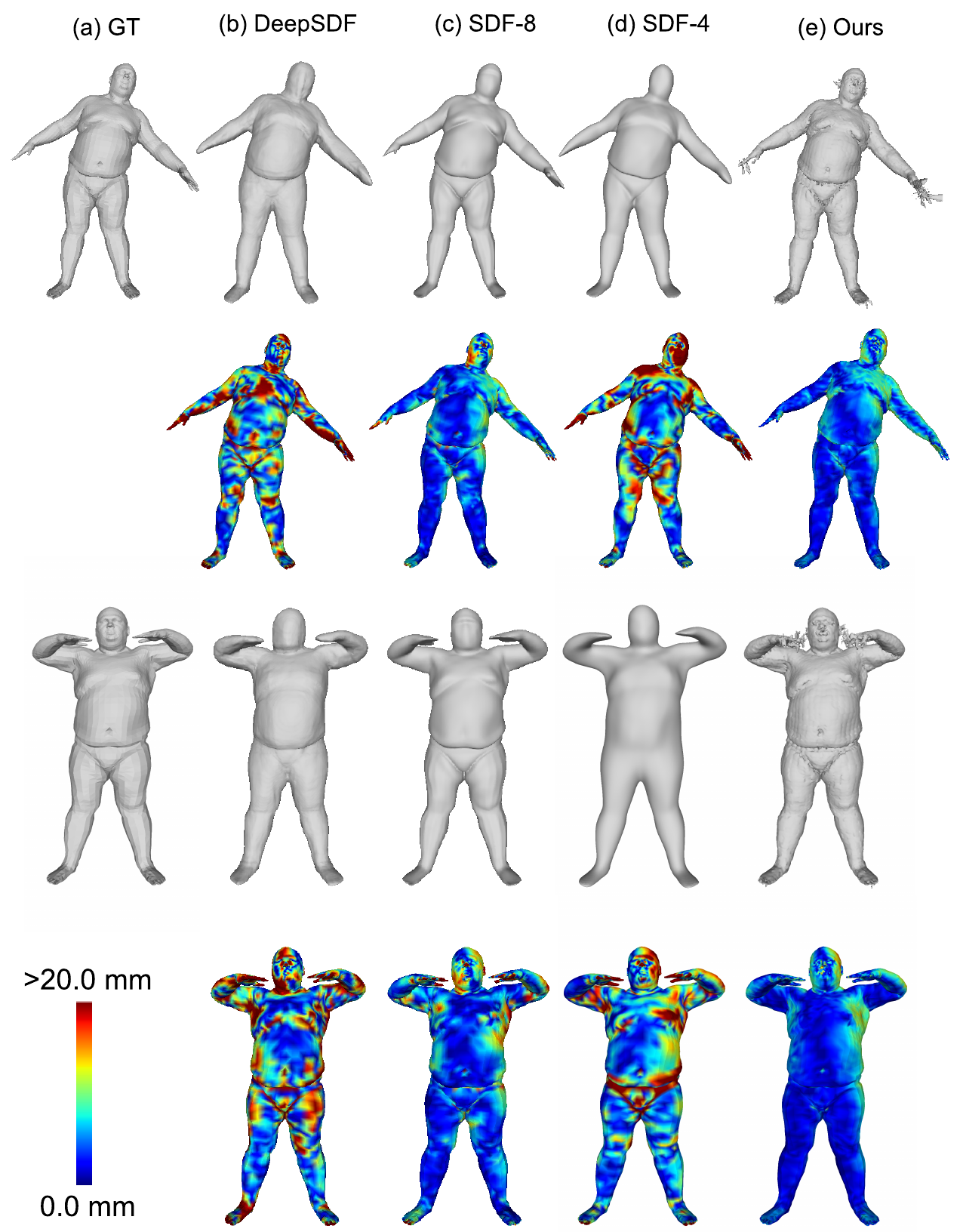}
\end{center}
\caption{Visualization of the per-vertex Euclidean error of the reconstructions.
GT means the ground truth shape, the model of Ours here used the LGCL-VC.}
\label{fig:compare2degeneration}
\end{figure}

We visualize two examples about the Euclidean error of each vertex shown in Fig~\ref{fig:compare2degeneration}.
It obviously shows that the small size of the DeepSDF network struggles to reconstruct the details, note that it almost loses the whole hands of the human.
In contrast, our LGCL model could keep more information in local regions though it causes more fluctuation.

\subsection{Ablation study}
\label{sec:ablation}

We perform ablative analysis experiments to evaluate the influence of our proposed geometric similarity loss $\mathcal{L}_{\mathrm{sim}}$.
It is controlled by adjusting its coefficient $\lambda_{\mathrm{sim}}$ to constrain the similarity between local latent codes with their neighbours.
As shown in Table~\ref{tab:lambda_influence},
the geometric similarity loss takes different influences on LGCL-VC and LGCL-Cheb.
Specifically, it tends to get better ED results with less constraint on similarities of local latent codes of LGCL-VC.

In contrast, one needs more similar local latent codes to decrease the errors of CD and HD since the large freedom of graph convolution kernels that used in LGCL-VC.
And for LGCL-Cheb, it implicitly has a stronger geometric constraint set by its ChebConv kernels.
Thus the extra geometric similar loss takes a little impact on the errors of CD and HD, but it should be patient to pick the adopted when you consider the ED errors.

\begin{table}
\begin{center}
\setlength{\tabcolsep}{3pt}{
\begin{tabular}{l|c|c|c|c|c|c}
\hline
\multirow{2}{*}{Model} &
\multirow{2}{*}{$\lambda_{\mathrm{sim}}$} &
\multirow{2}{*}{CD} &
\multirow{2}{*}{HD} & \multicolumn{3}{c}{ED} \\
 & & & & Mean & Std & Median \\
\hline
 \multirow{3}{*}{LGCL-VC}
 & 0.1 & 2.66 & 105.59 & 2.45 & 1.72 & 2.16 \\
 & 1.0 & 1.55 & 71.67 & 3.35 & 2.44 & 2.92 \\
 & 10.0 & 1.45 & 66.05 & 4.07 & 2.97 & 3.56 \\
 \hline
 \multirow{3}{*}{LGCL-Cheb}
 & 0.1 & 2.43 & 108.65 & 2.72 & 1.91 & 2.39 \\
 & 1.0 & 2.56 & 108.58 & 2.51 & 1.87 & 2.13 \\
 & 10.0 & 2.74 & 108.69 & 4.44 & 3.11 & 3.94 \\
\hline
\end{tabular}}
\end{center}
\caption{Influence of geometric similarity loss, all results are shown in millimetres.}
\label{tab:lambda_influence}
\end{table}

\begin{figure}[t]
\begin{center}
\includegraphics[width=0.9\linewidth]{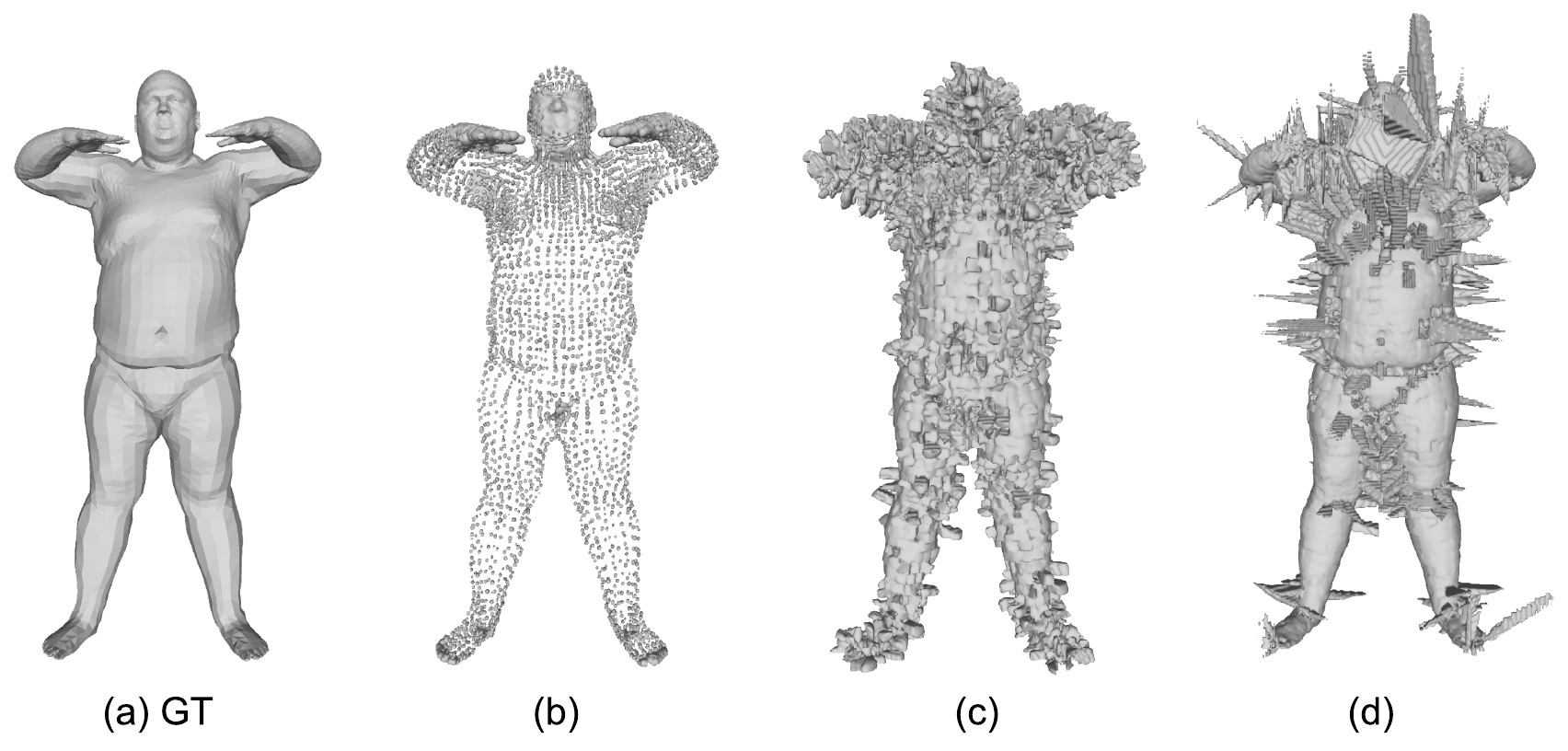}
\end{center}
\caption{Failed results of different methods for getting local latent codes.
(a) is the ground truth~(GT); 
(b) uses G2L but with the same initial input of global latent code for each graph vertex;
(c) uses a pyramid G2L to get local latent codes;
(d) does not include $\mathcal{L}_{\mathrm{sim}}$.
Please check more details in Sec~\ref{sec:ablation}.}
\label{fig:failed_results}
\end{figure}

We found some interesting explorations on the influence of different methods for getting local latent codes, as shown in Fig~\ref{fig:failed_results}.
In our LGCL-based methods, we split the global code $\mathrm{z}^G$ evenly into parts, which is equal to  the number of  the graph vertices, and then align these parts to different vertices.

However, for the result in Fig~\ref{fig:failed_results}~(b), we directly align the same initial input, which is the global latent code, to each vertex of the G2L.
Since each local latent code has the same initial value, it provides a similarity constraint implicitly between them. 
Then we do not introduce the $\mathcal{L}_{\mathrm{sim}}$.
In this case, each local region of reconstruction tends to shrink to the same type of mini polyhedron.
We attribute this degeneration of modelling local SDFs to the over-constraint on the similarity among the local latent codes, which introduces a limitation to geometric deformation of them.
We also show the results without the usage of $\mathcal{L}_{\mathrm{sim}}$ in our LGCL-based methods in Fig~\ref{fig:failed_results}~(d).
It is obvious to see the dramatic vibrations in some local regions.
We argue that it is caused by insufficient training of some local latent codes in corresponding local regions.

Furthermore, as shown in Fig~\ref{fig:failed_results}~(c), the result looks like fall in between (b) and (d), which has both similar polyhedron and vibrations that exist in local regions.
We got this result by modifying the G2L network as a pyramid structure as in the COMA~\cite{ranjan2018generating} decoder.
The modification changes the graph structure of each graph convolution layer with pooling layer.

The pyramid structure provides a cluster mapping from top to bottom in this G2L variant.
In other words, the mentioned modification adds an extra similarity constraint on local latent codes within a more large geometric range.
However, the approach is also limited to deform the local latent codes and leads to a compromise result compared to (b), (d) and our major results in Fig~\ref{fig:compare2degeneration}.

\section{Conclusions}

In this work, we propose the LGCL method which is based on  a  new architecture for the local geometry learning handling. 
The idea is to perform the learning regression process directly in the latent code space. Consequently, our approach makes general GNNs more flexible, compact, and simple in realization. 
The experimental results show that our method considerably outperforms baselines DeepSDFs both in accuracy and model size. 
We think that our architecture is novel, promising, and can be further improved in  future work.  

\section{Acknowledgements}

This work has been supported by the Spanish project TIN2015-65464-R, TIN2016-79717-R  (MINECO/FEDER) and the COST Action IC1307 iV\&L Net (European Network on Integrating Vision and Language), supported by COST (European Cooperation in Science and Technology).
We acknowledge the CERCA Programme of Generalitat de Catalunya.
Shun Yao acknowledges the Chinese Scholarship Council (CSC) grant No.201906290108.
We also acknowledge the generous GPU support from Nvidia.

{\small
\bibliographystyle{ieee_fullname}
\bibliography{main}
}

\end{document}